# Multi-PCA based Fault Detection Model Combined with Prior knowledge of HVAC

Ziming Liu[a]  Xiaobo Liu[a,b]

[a]College of energy and power engineering, Changsha University of Science and Technology, Changsha, 410014, China
[b]College of Civil Engineering, Hunan University, Changsha, 410082, China

**Abstract**

The traditional PCA fault detection methods completely depend on the training data. The prior knowledge such as the physical principle of the system has not been taken into account. In this paper, we propose a new multi-PCA fault detection model combined with prior knowledge. This new model can adapt to the variable operating conditions of the central air conditioning system, and it can detect small deviation faults of sensors and significantly shorten the time delay of detecting drift faults. We also conducted enough ablation experiments to demonstrate that our model is more robust and efficient.

## 1. Introduction

Fault detection and diagnosis(FDD) is important for a wide range of scientific and industrial processes, including central air-conditioning system(ACS). The central air-conditioning system is a complex coupling system, whose control system is regulated by negative feedback. Sensor is an important component in the ACS, and plays a key role in control system. However, sensor fault may mislead control system, resulting in wrong control commands, deviated operating conditions, waste of energy and even damage to equipment. So the sensor fault of ACS is the research object of the paper.

The fault detection method based on principal element analysis (PCA) is one of the main data-driven methods in FDD. Previous research has established that the method based is feasible in the field of HCAC[1-3]. Wang and Xiao established two PCA models based on the heat balance and pressure-flow balance to reduance the effect of the system non-linearity and enhangce the robustness[3]. Miller, Wang et al. proposed a multi-PCA modeling technique to realize monitory and fault detection in Rapid Thermal Anneal (RTA) process[4]. Turner, Staino et al. introduces a data-driven automated building HVAC fault detection method that uses a recursive least-squares model approach[5]. Du, Fan et al. developed the combined neural networks, integrating the basic neural network and auxiliary neural network, to detect the fault in the air handling unit(AHU)[6]. Schein, Bushby et al. used a rule-based, system-level fault detection and diagnostic (FDD) method to detect the faults of HVAC systems[7]. The research to date has tended to focus on knowledge-based methods or data-drived methods, but failed to address an association between knowledge-based methods and data-drived methods. Although some research has been carried out on PCA fault detection method in HVAC system, no studies have been found which combine PCA fault detection method with prior knowledge. Throughout this paper, the term 'prior knowledge' will refer to the rules or knowledges which are easyly obtained and related to HVAC system, such as climate change or crowd activity.

There is still uncertainty, however, whether PCA fault detection method is feasible in the multiple operating conditions of the central air conditioning system. This paper highlights the importance of prior knowledge and explores the ways applying PCA method to the non-linear and time dependent processes. There are two primary aims of this study: 1. To detect minor faults of HVAC system. 2. To adapt the effects of the non-linearity and enhance the robustness of method.

To realize the aims, multi-PCA fault detection method combined with prior knowledge is proposed in this paper. During offline modeling, a series of PCA submodel is built with the data of new equpiment running for one or two years, the number of submodel is determined by 'prior knowledge' of HVAC system, each sub-model describes a unique stable condition, all of the submodels are stroed in database. During online monitoring, an new observation is matched to the most matching operating condition in the database, and then it is monitored online by the PCA submodel corresponding to the condition. The study was conducted in the form of a computer simulation, with data being gathered via the dataset of the Central Air Conditioning System for the 5th Teddy Cup Data Mining Competition (Jointly presented by Guangzhou Teddy Intelligent Technology Co. and the Advanced Digital Science Center, University of Illinois, USA).

This work will generate fresh insight into data-drived FDD algorithm, the integration of prior knowledge can speed up the modeling and reduce the modeling cost. And Our model has a high failure detection rate and is robust. The remaining part of the paper proceeds as follows: first, we introduce trodtional PCA fault detection method, furthermore, multi-PCA fault detection method is proposed to adapt the nonlinear and time dependent process of HVAC system. Finally, the effectiveness and robustness of the new method are verified by experiment, using the dataset of an existing air-conditioning system.

## 2. Traditional PCA fault detection method

*2.1 Data processing*

Data preprocessing is a necessary preparation for fault detection and diagnosis algorithm, including data cleaning, data transformation, and data denoising. The quality of data processing directly determines the effectiveness of fault detection.

Data cleaning aims to analysize original data, correcting missing values and outliers. Missing values can be deleted directly in the detection of HVAC system, the outliers containing in the samples must be thrown, using Chauvenet's Criterion[8].

Data transformation is an essential step for PCA fault detection method. Because PCA modeling uses different sensor variables, including temperature, flow, pressure, etc., while the PCA algorithm requires dimensional uniformity, data normalization is often used to eliminate the influence of dimension. The formula is as follows:

$$z = \frac{x-\mu}{\sigma} \qquad (1)$$

Where x is the value of the variable, μ is the mean, and σ is the standard deviation.

For data denoising, wavelet denoising has been proved to be an effective method to improve the performance of algorithm[9].

*2.1 PCA model*

Assuming a data matrix under normal operating conditions $X_0 \in R^{n \times m}$ (where n is the number of samples, m is the number of variables), the normalized matrix X can be decomposed :

$$X = \hat{X} + E \qquad (2)$$

Suppose that the top l principal components of X reflect the information of the main feature $\hat{X}$, and the last (m-l) principal components reflect the information of the secondary feature E.

$$\hat{X} = \hat{T}\hat{P}^T = \sum_{i=1}^{l} t_i p_i^T \qquad (3)$$
$$E = \tilde{T}\tilde{P}^T = \sum_{i=l+1}^{m} t_i p_i^T \qquad (4)$$

Where l(l < m) is the number of principals of the model, $T \in R^{n \times l}$ is the score matrix, and $P \in R^{m \times l}$ is the load matrix.

As the model is established, the sample vector to be measured can be decomposed into two parts: **the model principal subspace PCS and the residual subspace RS**:

$$x = \hat{x} + \tilde{x} \qquad (5)$$

where $\hat{x} = Cx = \hat{P}\hat{P}^T x$ and $\tilde{x} = \tilde{C}x = \tilde{P}\tilde{P}^T x = (I - C)x$.

Under normal circumstances, $\hat{x}$(the projection of x in the PCS) mainly represents the normal value of the data, while $\tilde{x}$(the projection of x in the RS) is mainly represents the noise value of the data. When a fault occurs, the projection $\tilde{x}$ in the RS will increase significantly due to the fault, which is more sensitive than projection $\hat{x}$ in the PCS. Fault detection will be performed by both of them.

*2.3 Fault detection model based on PCA*

The PCA-based fault detection model is mainly based on three indicators. The detection indicators include the $T^2$ statistic that characterizes the principal subspace and the SPE statistic that characterizes the residual subspace. In addition, combined with the $T^2$ statistic and The SPE statistic, we can define a comprehensive indicator φ.

Hotelling's $T^2$ statistic measures the change of the sample vector in the principal space. So the $T^2$ statistic usually represents the change of the normal data. Therefore, the fault is usually detected by the SPE statistic instead of the $T^2$ statistic. The calculation process is as follows.

$$T^2 = x^T P \Lambda^{-1} P^T x \leq T_\alpha^2 \qquad (6)$$

Where $\Lambda = \text{diag}\{\lambda_1, \lambda_2, \lambda_3, \dots, \lambda_A\}$, $T_\alpha^2$ is the control limit of $T^2$ when the confidence is α. $T_\alpha^2$ can be calculated as follows[10]:

$$T_\alpha^2 = \frac{A(n^2-1)}{n(n-1)} F_{A,n-A;\alpha} \qquad (7)$$

Where $F_{A,n-A;\alpha}$ is an F-distribution value with A or (n-A) degrees of freedom and a confidence of α.

The SPE statistic measures the change of the projection of the sample vector in the residual space. So the SPE statistic represents the noise variation of the residual subspace. Because the fault is usually regarded as the noise of the normal data, the SPE statistic will be significantly improved when the fault occurs. The calculation process is as follows:

$$\text{SPE} = \|(I - PP^T)x\|^2 \leq \delta_\alpha^2 \qquad (8)$$

Where $\delta_\alpha^2$ is the control limit of SPE when the confidence is α. $\delta_\alpha^2$ can be calculated as follows[11, 12]:

$$\delta_\alpha^2 = \theta_1 \left( \frac{C_\alpha \sqrt{2\theta_2 h_0^2}}{\theta_1} + 1 + \frac{\theta_2 h_0(h_0-1)}{\theta_1^2} \right)^{\frac{1}{h_0}} \qquad (9)$$

Where $\theta_i = \sum_{j=A+1}^{m} \lambda_j^i$, $i = 1,2,3$. $h_0 = 1 - \frac{2\theta_1 \theta_3}{3\theta_1^2}$. $\lambda_j^i$ is the eigenvalue of the covariance matrix of X, C is the threshold with a confidence of α under the standard normal distribution.

Sometimes it is more convenient to use a combined index φ to monitor faults. So Yue and Qin propose a combined index φ combining SPE and $T^2$ to monitor faults[13]. In this paper, we use combined index φ to detect sensor fault, which can solve the inconsistency of the detection results between $T^2$ and SPE.

$$\varphi = \frac{\text{SPE}(x)}{\delta_\alpha^2} + \frac{T^2(x)}{T_\alpha^2} = x^T \Phi x \tag{10}$$

$$\Phi = \frac{P\Lambda^{-1}P^T}{T_\alpha^2} + \frac{I-PP^T}{\delta_\alpha^2} \tag{11}$$

Where $\Phi$ is symmetric and positive definite.

## 3 Multi-PCA fault detection method combining prior knowledge

The air conditioning system is a closed-loop system. When the system load changes, the control loops run in the fixed value adjustment mode. If the set value remains unchanged, The process parameters are maintained around the set value due to the closed-loop control. And these changes all satisfy the characteristics of normal distribution. However, the operating conditions of the air conditioning system are constantly changing, switching back and forth between "steady state - transition state - steady state". On the one hand, the data distribution of the central air conditioning system satisfies the conditions of the PCA fault detection method. On the other hand, the characteristics of the variable operating conditions make the single PCA fault detection model lose effectiveness. So, the method of the multi-PCA model is proposed to Solve the problem of variable operating conditions.

### 3.1 Basic ideas and model structure

Multi-PCA model has been proposed early, which mainly uses the clustering algorithm to realize the division of working conditions, and then establish different PCA submodel for different working conditions. In this paper we call it **mPCA1**, mPCA1 is a fully data driven multi-PCA model which has blindness and uncertainty. In the modeling process, the optimal model group must be established by continuously trying different cluster numbers K and cross-validation. So mPCA1 will become inefficient and costly to model[4].

In order to solve these problems, we further proposed a multi-PCA fault detection method combining prior knowledge which is called mPCA2 in this paper. The innovation of mPCA2 is that we added the simple physical knowledge of the central air conditioning system to the model, including outdoor conditions, indoor conditions and the operation of the system itself. These simple physical knowledge is prior knowledge.

**The basic idea** of the multi-PCA model combined with prior knowledge is as follows: Firstly, the number of operating conditions of the modeling data (K) was determined by the prior knowledge, and then the modeling data of the same operating condition was collected by using the clustering algorithm. A corresponding PCA sub-model was established for the data of each working condition. When new sample data was generated, the new sample was matched with all sub-models in the database. If there was a corresponding working condition, the sub-model of the working condition can be used for fault detection.

We think that the complete operation data of the first 1-2 years of the new equipment can contain all the operating conditions that the equipment may have in the future. Because the sensor failure of the central air-conditioning system usually occurs after the device has run for 1-2 years. These data are used as modeling data to establish the multiple PCA model. The obtained PCA submodel group is stored in the database.

For the central air conditioning system, the operating conditions of the system are determined by the indoor conditions, the outdoor conditions and the conditions of the system itself. The research in this paper found that:

1. The central air conditioning system is generally equipped with multiple chillers and multiple pumps. The operating combination of different units roughly determines the operating conditions of the system.
2. There are significantly different operating rules for operation under summer conditions and operation under winter conditions, indicating that outdoor climate can be used to determine system operating conditions.
3. The intensity of the crowd inside the building and the outdoor conditions together determine the load of the system and determine the operating conditions of the system. For example, office buildings are typical buildings that are affected by the intensity of indoor crowd activity. During work hours, the density of people in the building is large and basically constant, and the density of people in the building during the break period is drastically reduced. Therefore, the working hours and rest time can correspond to two significantly different operating conditions.

In summary, we believe that the operation of the unit, the outdoor climate, and the intensity of the crowd inside the building can be used as prior knowledge to divide the system conditions.

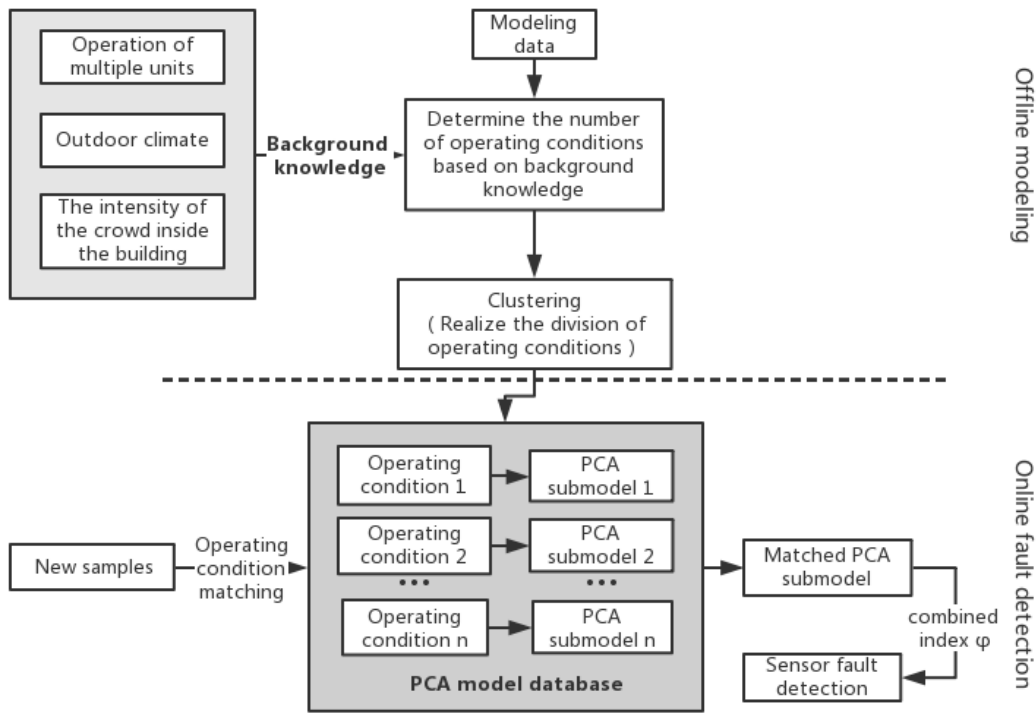

Fig. 1 Structure of fault detection model based on MPCA2

*3.2 Modeling steps*
*3.2.1 Modeling steps for the MPCA2 model:*
  1. Obtain raw data $X_0(N_0 \times M)$ and perform data preprocessing.
  2. The prior knowledge of the system determines that there are K operating conditions in the system. The clustering model divides the original data into K working conditions $X_1(N_1 \times M), X_2(N_2 \times M), \ldots, X_K(N_K \times M)$, where $N_0 = N_1 + N_2 + \cdots + N_K$, M is the number of characteristic variables of the system.

3. Perform principal component analysis on process data $X_1(N_1 \times M), X_2(N_2 \times M), \ldots, X_K(N_K \times M)$, and calculate the corresponding load matrix $P_1(M \times L_1), P_2(M \times L_2), \ldots, P_K(M \times L_K)$.

4. Calculate the control limits of $T^2$ statistic, SPE statistic and comprehensive indicator according to formula (7) (9) (10).

*3.2.2 Steps for fault online monitoring:*

1. Data preprocessing of the new sample data $x_0(1 \times M)$.

2. Match the new data $x_0$ to the closest working condition i and calculate the corresponding load matrix.

3). Calculate the $T^2$ statistic, SPE statistic and the comprehensive index $\varphi$ of the new data according to formulas (6) (8) (10), and compare them with the corresponding control limits. If the comprehensive indicator $\varphi$ exceed its control limit, which indicates that there is fault happened, otherwise it is normal sample.

**4 Fault detection test in a water-cooled central air-conditioning system**

In order to verify the effectiveness and robustness of the new method, the experiment was carried out using real-life data from the equipment. The experimental data came from an existing water-cooled central air-conditioning system in Singapore and was gathered via the dataset for the 5th Teddy Cup Data Mining Competition (Jointly presented by Guangzhou Teddy Intelligent Technology Co. and the Advanced Digital Science Center, University of Illinois, USA). In order to train multiple PCA models, normal data of 4 days in October and 4 days in November were used to train the PCA model, and the data of 2 days in November was used to generate simulated faults to verify whether the new method has significantly improved the fault detection effect of sensor's deviation fault and drift fault. All the data were sampled at an interval of 1 min.

To simplify the model, The characteristic variables of the model include 6 process variables: 1.the temperature of the water flowing into the cooling device, 2.the temperature of the water flowing out of the cooling device, 3.the flow rate of the cooling cycle, 4.the temperature of the water flowing into the condensing device, 5.the temperature of the water flowing out of the condensing device, and 6.the flow rate of the condensing cycle. All of them can be found in Figure 2. After removing the outliers and transient data, 2662 samples were used to construct the training matrix $(2662 \times 6)$.

The verification process is divided into 2 steps. First, a single PCA model is used to detect sensor failures, in which only one PCA model is established. Secondly, the new proposed fault detection method using the multi-PCA model combined with prior knowledge is used to detect sensor faults, in which multiple PCA sub-models are established.

*4.1 analisis of prior knowledge*

The system consists of 3 chillers (cooling devices), 4 chilled water pumps, 3 condensate pumps, and 2 cooling towers, the structure of system is shown as the Figure 2. In general, all equipments of a central air conditioning system will not operate at the same time, and most of the time there will be different combinations of unit operations. For the test data, we can easily divide it into three different running states according to the operation of the units.

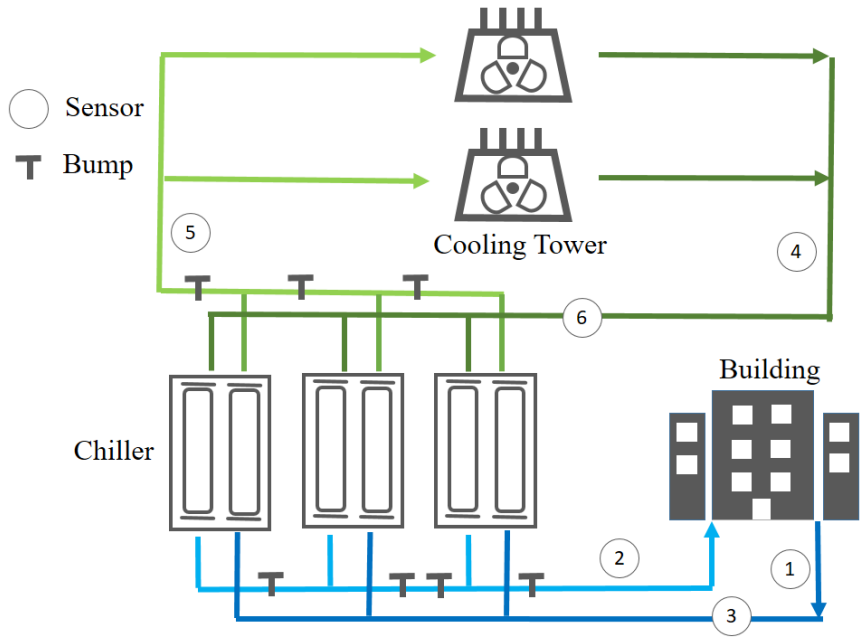

*Fig 2. Construction of the system*

In Singapore, the annual temperature does not change much, and the average temperature is around 23-31 degrees Celsius. The rainy season in Singapore is from December to March, the weather is relatively humid, with an average temperature of around 23-24 degrees. The dry season is between June and September, when the southwest monsoon is blown and the weather is relatively dry. The other months belong to the Monsoon alternating month. So we can say that outdoor climate of sigapore includes rainy season, dry season and monsoon alternating month.

For the intensity of the crowd inside the building, to simplify the model, it is divided into two types of operating conditions according to the difference of working time and rest time.

All of the prior knowledge mentioned above is easy to obtain by BMS of building and can be recognized by computer programs. In this experiment, the modeling data containing 2262 samples was divided into three sample sets after prior knowledge analysis, as shown in the following table.

*Table 1. the result of the prior knowledge analysis*

| Operating conditions | Operation of multiple units | Outdoor climate | intensity of the crowd inside the building |
|---|---|---|---|
| No.1 | Runing State 1 (2 chillers,5 bumps) | Monsoon alternating month | Working time |
| No.2 | Runing State 2 (2 chillers,4 bumps) | Monsoon alternating month | Working time |
| No.3 | Runing State 3 (1 chillers,2 bumps) | Monsoon alternating month | Rest time |

*4.2 sensor's deviation fault*

In order to compare the effect of the two methods on the deviation fault of the temperature sensor. We simulated the fault with the data of 2 days in November, which s an 11% deviation fault (A subtle deviation about 1 °C) on sensor No.1.

At the first step of the validation process, the traditional PCA fault detection method was used to monitor the sensors of system. The result of fault detection was shown in Figure 3.

There are 1032 samples for testing, and the failure detection rate is 7.36%. The traditional PCA fault detection method has failed to find the deviation fault because the combined index of most samples do not exceed threshold but these samples are fault samples as we have known. The result shows that the traditional PCA fault detection method fails under multiple operating conditions, and it cannot detect subtle sensor deviation faults.

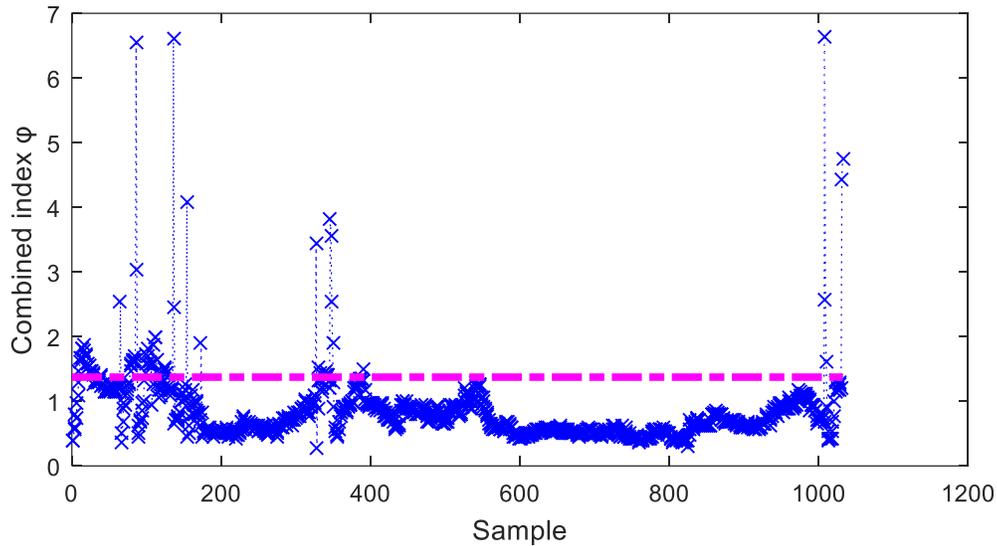

*Fig. 3 sensor's deviation Fault detection results of the traditional PCA model*

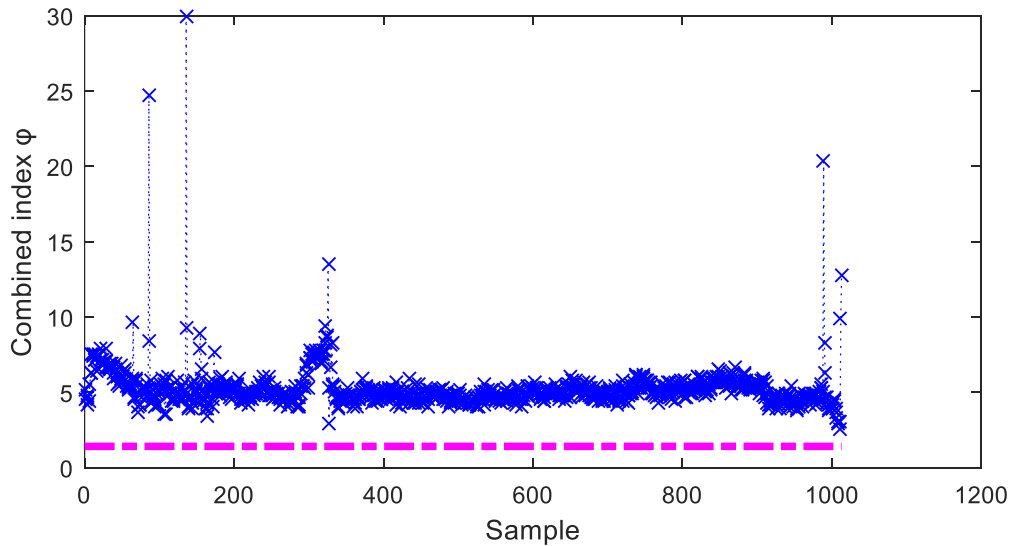

*Fig. 4 sensor's deviation Fault detection results of new method*

At the second step of the validation process, new multi-PCA fault detection model is used to monitor the sensor. There were 1032 failure samples for testing, and the most samples which is 1013 samples is matched into PCA submodel No.3. The plot of combined index of submodel No. 3 is shown in Figure 4. There is 1013(100% of the valid samples and 98.16% of the total test samples) samples was detected to be abnormal. **Therefore, the newly proposed fault detection method significantly improves the fault detection effect, and successfully applies the PCA fault detection model to the central air conditioning system in variable working conditions.**

In addition, in the modeling process of multiple PCA models, there are three PCA sub-models stored in the database, the details are shown in Table 2.

Table 2 Details of PCA submodles

| PCA submodle | $T^2$ threshold | SPE threshold | Combined index φ |
|---|---|---|---|
| No. 1 | 6.1000 | 1.0423 | 1.3551 |
| No. 2 | 9.5279 | 1.3031 | 1.4009 |
| No. 3 | 9.6245 | 0.1389 | 1.4113 |

*4.3 sensor's drift fault*

To study the effect of new method on the drift fault of sensor, we simulated the fault with the data of 2 days in November, which is drift fault of 2.5°C on sensor No.1. Two methods including traditional PCA model and multiple-PCA model combined with prior knowledge were tested.

There are 1032 drift fault samples to be tested. The result of the traditional PCA model is shown in Figure 5, and there are 111 samples(about 10.76% of all samples) are detected to be abnormal, which is too little to prove the failure. For the multiple PCA method, most of the samples including 800 valid samples are matched into PCA submodel No.3, and there are 715 samples(about 89.38% of valid samples and 69.28% of total test samples) detected to be abnormal, which is shown in Figure 6. We can find that the detection efficiency of the drift fault is lower than that of the deviation fault, whether it is the traditional PCA method or the multi-PCA method, although the drift fault imposes a larger fault. Because the drift fault is slowly increasing, at the beginning, the fault is almost non-existent, and the fault becomes more apparent over time.

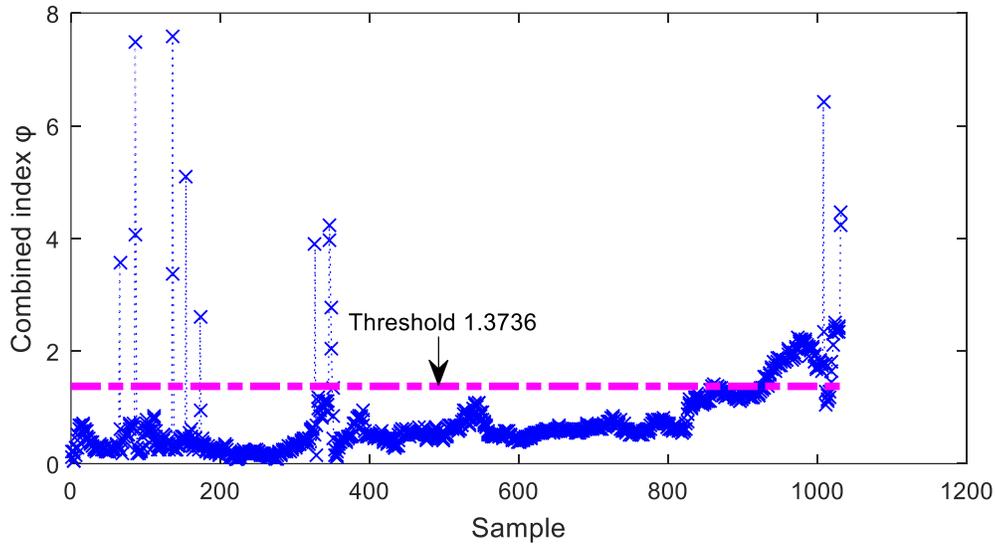

*Fig. 5 sensor's drift fault detection results of traditional PCA model*

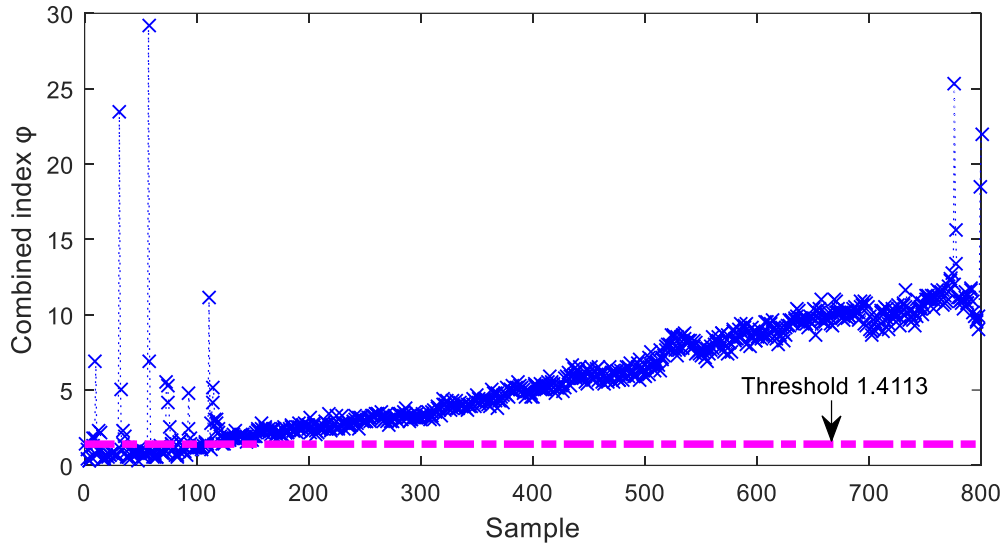
*Fig. 6 sensor's drift fault detection results of the new method*

Observing the trend of the two results, it can be found that the combined index of the samples shows a gradual upward trend. Since the 900th sample in the traditional PCA model, the combined index exceeded the threshold and remained above the threshold. A drift fault was detected during the last tenth of a time period when the system failure had occurred for two days. Using the new proposed method, the combined index has exceeded the threshold and continued to rise since the 220th sample. Drift failure can be detected on the first day of its occurrence. In general, the detection of drift faults must have a time delay. Applying the new method proposed in this paper can significantly reduce this time delay and detect faults earlier.

## 4. Conclusion

Subtle sensor failures are often difficult to detect, especially in variable operating conditions, where subtle faults are more difficult to detect. Based on the traditional PCA fault detection technology, this paper proposes a multi-PCA fault detection method based on prior knowledge. The two experiments designed prove that the new method has better sensitivity and fault detection capability, which is embodied in two aspects: First, the multi-PCA fault detection method can detect very subtle deviation faults. Second, the new method significantly reduces the time delay for detecting drift faults. Deviation faults and drift faults are the two most common sensor's fault and difficult to be detected.

The core problem of the multi-PCA fault detection method is to solve the division problem of the operating conditions. In general, the operational data of the new device for the first 1-2 years is used for modeling, and the database is established to store all PCA sub-models. Most of the past researches use the model that relies entirely on data-driven. we proposes a multi-PCA model that combines the simple prior knowledge of central air-conditioning systems to compensate for the blindness and uncertainty of complete data-driven algorithms. Experiments show that the efficiency and robustness of the model are improved, and the fault detection effect is better than traditional PCA fault detection method.